%% file: arxiv.tex
\documentclass[letterpaper, 10 pt, conference]{ieeeconf}  

\IEEEoverridecommandlockouts                              

\usepackage{graphics} 
\usepackage{times} 
\usepackage{amsmath} 
\usepackage{amssymb}  
\usepackage{bm}

\usepackage{subcaption}
\usepackage{graphicx}
\usepackage{xcolor}
\usepackage{booktabs}
\usepackage{xspace}
\usepackage{url}
\usepackage{algorithmic}
\usepackage[linesnumbered,ruled,vlined]{algorithm2e}
\usepackage[export]{adjustbox}
\usepackage{chngpage}
\usepackage[section]{placeins}
\usepackage{bm}
\usepackage{multirow}
\usepackage{tabularx}
\usepackage{dsfont}
\usepackage{physics}
\usepackage[pagebackref=true,breaklinks=true,colorlinks,bookmarks=false]{hyperref}

\DeclareMathOperator*{\argmax}{arg\,max}

\graphicspath{ {./images/} }

\def\ourmethod{{MOMA-Force}\xspace}

\title{\LARGE \bf
MOMA-Force: Visual-Force Imitation for Real-World \\ Mobile Manipulation
}

\author{Taozheng Yang$^{*}$, Ya Jing$^{*}$, Hongtao Wu$^{*}$, Jiafeng Xu$^{*}$,  Kuankuan Sima$^{\dagger}$, \\Guangzeng Chen, Qie Sima$^{\dagger}$, Tao Kong$^{\ddagger}$ \\ 
ByteDance Research \\
\url{https://visual-force-imitation.github.io}
\thanks{
$^{*}$Equal contribution. $^{\dagger}$Work done during an internship.}
\thanks{$^{\ddagger}$Corresponding author: Tao Kong (\texttt{kongtao@bytedance.com}).
} 
}

\begin{document}

\maketitle
\thispagestyle{empty}
\pagestyle{empty}

\begin{abstract}
\input{00abstract}
\end{abstract}

\IEEEpeerreviewmaketitle

\section{Introduction}
\input{10intro}\label{sec:intro}

\section{Related Work}
\input{20related}

\section{Method}
\input{30method}\label{sec:method}

\section{Experiments}\label{sec:expr}
\input{40experiment}


\section{Conclusion} \label{sec:conclusion}
\input{60conclusion}


\bibliographystyle{IEEEtran}
\bibliography{references}

\clearpage
\appendix
\input{appendix}

\end{document}

%% file: 00abstract.tex
In this paper, we present a novel method for mobile manipulators to perform multiple contact-rich manipulation tasks. 
While learning-based methods have the potential to generate actions in an end-to-end manner, they often suffer from insufficient action accuracy and robustness against noise. 
On the other hand, classical control-based methods can enhance system robustness, but at the cost of extensive parameter tuning.
To address these challenges, we present \ourmethod, a visual-force imitation method that seamlessly combines representation learning for perception, imitation learning for complex motion generation, and admittance whole-body control for system robustness and controllability.
\ourmethod enables a mobile manipulator to learn multiple complex contact-rich tasks with high success rates and small contact forces. 
In a real household setting, our method outperforms baseline methods in terms of task success rates.
Moreover, our method achieves smaller contact forces and smaller force variances compared to baseline methods without force imitation.
Overall, we offer a promising approach for efficient and robust mobile manipulation in the real world.
Videos and more details can be found on \url{https://visual-force-imitation.github.io}.

%% file: 10intro.tex
\begin{figure}[t]
\centering
\includegraphics[width=\linewidth]{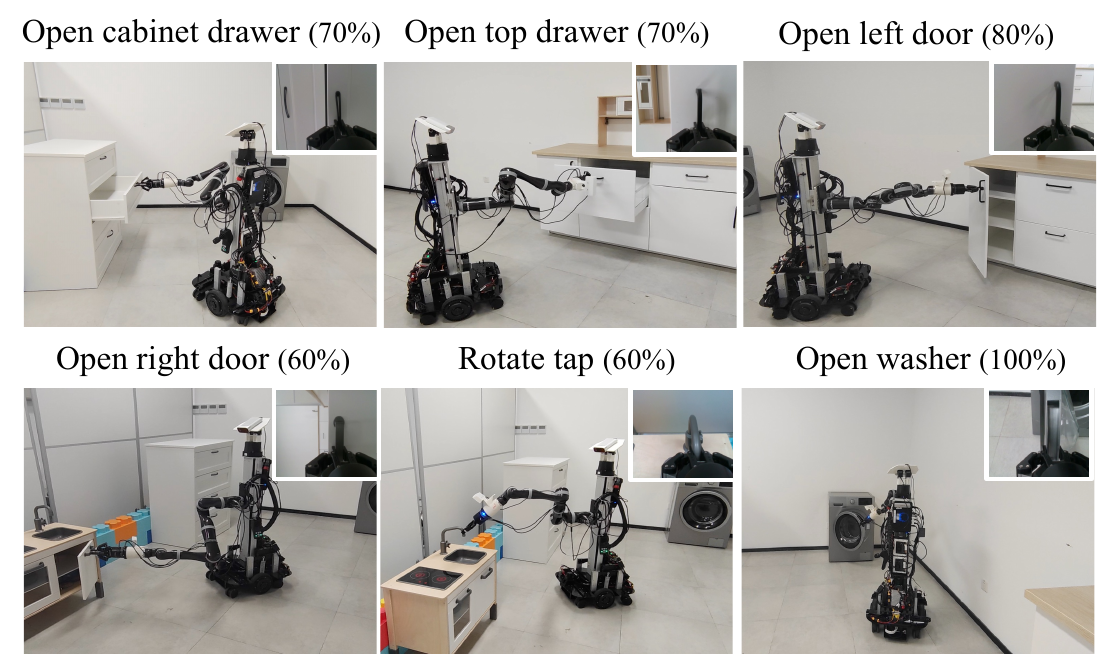}
\caption{\textbf{Overview.} We perform six real-world contact-rich mobile manipulation tasks. The success rates of our method are shown at the top of each image. 
\label{fig:illu}}
\vspace{-0.7cm}
\end{figure}

Mobile manipulation combines two fundamental robot capabilities: mobility and manipulation. 
These two capabilities substantially extend robot applications in the real world compared to static manipulation~\cite{wong2022error}. 
For example, mobile manipulation enables robots to complete tasks involving manipulations with large workspaces (\textit{e.g.}, opening closet doors).
However, mobile manipulation poses significant challenges when it comes to real-world tasks.
The challenges are mainly twofold.
First, uncertainties caused by localization and control can lead to potential safety issues, especially in contact-rich tasks.
Second, the high-dimensional configuration space makes motion generation and control complex.

Early work assumed prior knowledge of the scene and explored path planning \cite{carriker1991path} and task and motion planning (TAMP) \cite{wolfe2010combined} for mobile manipulators.
Whole-body control algorithms have been studied to solve high-dimensional optimization problems \cite{mittal2021articulated, minniti2019whole, pankert2020perceptive}.
However, the environment is also simplified to ease the optimization complexity.
Data-driven methods such as imitation learning \cite{nair2022r3m,radosavovicreal}
and reinforcement learning \cite{fudeep} are becoming popular as they enable robots to learn directly from visual observation.
Recent works extend the large-scale pre-training paradigm from computer vision and natural language processing to static manipulation~\cite{nair2022r3m,radosavovicreal, radford2021learning} and navigation~\cite{ramrakhya2022habitat}.
They show promising results in various real robot tasks.
Compared to exclusively learning from expensive robot data, this paradigm makes use of existing large-scale dataset~\cite{deng2009imagenet, grauman2022ego4d} to pre-train a visual representation model which is able to learn useful visual properties of the world for efficient policy learning.
However, these methods are rarely designed for contact-rich mobile manipulation tasks. 
And visual imitation learning generally suffers from the ``last-centimeter" error~\cite{bcz}. 
This problem becomes non-negligible in many real-world contact-rich tasks as it may lead to severe mechanical damage to the environment and robots.

In this paper,  we argue that wrench $\mathcal{F}=[m, f] \in \mathbb{R}^{6}$ ($m$ is the torque and $f$ is the force) is critical in terms of task completion and safety. 
To this end, we propose a novel visual-force imitation method for solving real-world contact-rich mobile manipulation tasks.
In contrast to previous work which only imitates kinematic actions, we extend a state-of-the-art visual imitation method \cite{pari2021surprising} to support both kinematic action and force imitation.
This enables us to tackle the uncertainties caused by localization and control in mobile manipulation, and thus ensures safer robot execution.
Moreover, we leverage whole-body control (WBC)~\cite{pankert2020perceptive, sleiman2021constraint} to solve the high-dimensional control problem of mobile manipulators.
This allows the robot to track the trajectory generated by the predicted action while regulating the wrench to imitate the expert wrench.
We perform multi-task learning on six household contact-rich tasks (Fig. \ref{fig:illu}).
Results show that our method outperforms several baseline methods without force imitation on task success rate.
In particular, our method achieves an average success rate of 73.3\% on the six tasks, while the best baseline method achieves 45.0\%.
Additionally, with force imitation, the average absolute contact wrench and the average wrench variance decrease, indicating a safer and more stable contact between the robot and the object.

The key contributions are summarized as follows: 
\begin{itemize}
    \item We address the challenging problems caused by uncertainty and high-dimensional kinematics in real-world mobile manipulation tasks in an imitation learning context, and propose an effective visual-force imitation method to solve real-world mobile manipulation tasks.
    \item We present systematic real robot experiments on six contact-rich mobile manipulation tasks and showcase the performance of the proposed system on these challenging tasks. 
\end{itemize}

%% file: 20related.tex
\subsection{Mobile Manipulation}
Previous work explored using task and motion planning (TAMP) to plan and control mobile manipulators~\cite{wolfe2010combined}.
Another line of methods leverages a move-and-act pipeline~\cite{roa2021mobile}.
However, it is often challenging to find viable kinematic solutions due to constraints from robots and/or environments.
Alternatively, whole-body control \cite{haviland2021neo, haviland2022holistic} are able to deal with complex constraints arise from mechanical limitations, manipulability or dynamic obstacle avoidance requirements.
Learning-based methods are adopted in mobile manipulation to enable robots with visual feedback. 
In particular, imitation learning methods try to mimic actions from human demonstrations~\cite{welschehold2017learning,kazhoyan2020learning,brohan2022rt}.
Additionally, force information must be carefully considered to improve task completion and safety~\cite{roa2021mobile}. 

\subsection{Robot Learning from Demonstrations}
Learning from demonstrations is a sample-efficient and practical approach for training robots to perform complex manipulation skills.
A popular line of work leverages behavior cloning (BC) to efficiently learn from human demonstrations \cite{torabi2018behavioral,radosavovicreal,young2021visual, bcz, florence2022implicit, shridhar2023perceiver}.
A recent work trains a transformer to learn multiple tasks with a dataset containing $\sim$130k episodes~\cite{brohan2022rt}. 
Additionally, there has been growing interest in learning general representations for robot manipulation based on pre-trained models \cite{nair2022r3m,radosavovicreal,gupta2022maskvit,xiao2022masked}.
Inspired by these methods, we use a pre-trained model to extract visual representations in this paper. 
The most related work is a non-parametric method that retrieves expert actions by comparing the representation of the observation image captured in the rollout with those of the observations in the expert trajectories~\cite{pari2021surprising}. 
Our method differs in that we imitate both kinematic actions and wrenches while this work only imitates kinematic actions. 

\subsection{Force Compliant Manipulation}
In classical control community, 
several early works on force compliance control have been developed. 
Specifically, \cite{Hogan1987StableEO} and \cite{Ott5509861} provide implementations of impedance and/or admittance control for robot manipulation based on wrench feedback. 
Manipulation tasks with articulated constraints are very common in household environments, \textit{e.g.}, opening doors and drawers.
Force compliance control is generally used to provide the required compliance for tackling these tasks~\cite{meeussen2010autonomous}.
Force compliance is also essential for dexterous in-hand manipulation and teleoperation \cite{zeng2021learning}. 
Inspired by the previous work, we believe force imitation is able to enhance the performance of mobile manipulators through improved action accuracy and safer environment interaction.

%% file: 30method.tex
\begin{figure*}[ht]
\centering
\includegraphics[width=1.5\columnwidth,trim=0 0 0 0,clip]{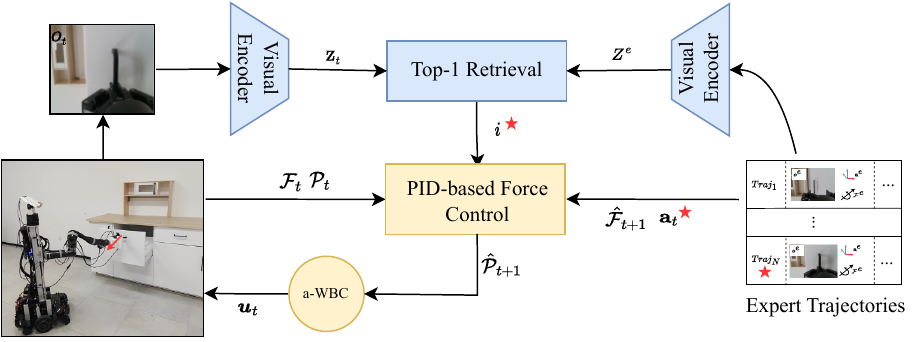}
\caption{\textbf{\ourmethod} The observation images of the expert data are converted to representation vectors by a visual encoder. In the rollout, the observation image is converted to a representation vector with the same visual encoder. The action and target wrench are predicted by retrieving the action and wrench of the expert data with top-1 similarity of the representation vectors. We use admittance whole-body control (a-WBC) to control the robot.}
\label{fig:method}
\vspace{-0.5cm}
\end{figure*}

We consider contact-rich mobile manipulation tasks where a robot interacts with the environment or an object within the environment. 
Besides manipulation, such tasks also necessitate the integration of mobility due to large workspace.
Our method mainly consists of two parts: 1) an action-wrench prediction module and 2) an admittance whole-body control module (Fig. \ref{fig:method}).
At each time step $t$, the action-wrench prediction module takes as input the RGB image $\mathbf{o}_{t}$ captured from an arm-mounted camera and generates a kinematic action $\mathbf{a}_{t} \in SE(3)$, a gripper action $\mathbf{g}_{t} \in \{-1, 0, 1\}$, a target wrench for the next time step $\widehat{\mathcal{F}}_{t+1} \in \mathbb{R}^{6}$ and a terminate flag $\mathcal{T}_{t} \in \{0, 1\}$ which indicates rollout termination.
If $\mathbf{g}_{t}=0$ and $\mathcal{T}_{t} = 0$, the control module takes as input $\mathbf{a}_{t}$ and $\widehat{\mathcal{F}}_{t+1}$ and controls the robot to track a target pose generated by $\mathbf{a}_{t}$ while regulating the contact wrench to imitate $\widehat{\mathcal{F}}_{t+1}$.
If $\mathbf{g}_{t}=1$ (or  $\mathbf{g}_{t}=-1$), the robot opens (or closes) the gripper.
If $\mathcal{T}_{t} = 1$, the rollout terminates.
We assume access to an expert dataset $D$ containing expert trajectories of \textit{multiple} tasks.
Each trajectory consists of multiple frames $d$ captured at different time steps.
Each frame is a tuple $d = \{\mathbf{o}^{e}, \mathbf{a}^{e}, \mathbf{g}^{e}, \mathcal{F}^{e}, \mathcal{T}^{e}\}$ where $\mathbf{o}^{e}$, $\mathbf{a}^{e}$, $\mathbf{g}^{e}$, $\mathcal{F}^{e}$, and $\mathcal{T}^{e}$ are the observation image, kinematic action, gripper action, wrench, and terminate flag, respectively.
We use superscript $e$ on variables to indicate that they are from the expert dataset $D$.
Alg. \ref{alg:method} shows the algorithm of our method.

\begin{algorithm}[t]
\DontPrintSemicolon
\KwIn{\\
\quad $D$ - Expert dataset\\
\quad $i_{\mathrm{max}}$ - Maximum rollout iteration}
$Z^{e} \gets \emptyset$\\
\For{$d \gets D$}{
    $(\mathbf{o}^{e}, \mathbf{a}^{e}, \mathcal{F}^{e}) \gets d$\\
    $\mathbf{z}^{e} \gets$ VisualEncoder$(\mathbf{o}^{e})$\\
    $Z^{e} \gets Z^{e} \cup \{\mathbf{z}^{e}\}$
}
$i \gets 0$\\
\While{$i < i_{\rm{max}}$}
{
    $i \gets i+1$\\
    $(\mathbf{o}_{t}, \mathcal{F}_{t}, \mathcal{P}_{t}) \gets$ getCurrentState$()$\\
    $\mathbf{z}_t \gets$ VisualEncoder$(\mathbf{o}_{t})$\\
    $(\mathbf{a}_{t}, \widehat{\mathcal{F}}_{t+1}, \mathcal{T}_{t}) \gets$ Retrieve$(\mathbf{z}_{t}, Z^{e})$\\
    \If{$\mathcal{T}_{t} = 1$}
    {
        \textbf{return}
    }

    $\widehat{\mathcal{P}}_{t+1} \gets \mathcal{P}_{t} \circ \mathbf{a}_{t}$\\

    $\Delta \mathcal{P}_{t+1} \gets$ ForceControl$(\mathcal{P}_{t}, \widehat{\mathcal{P}}_{t+1}, \mathcal{F}_{t}, \widehat{\mathcal{F}}_{t+1})$\\
    $\widehat{\mathcal{P}}_{t+1} \gets \widehat{\mathcal{P}}_{t+1} \circ \Delta \mathcal{P}_{t+1}$

    a-WBC$(\widehat{\mathcal{P}}_{t+1})$
}
\textbf{return}
\caption{VisualForceImitation}
\label{alg:method}
\end{algorithm}

\subsection{Action-Wrench Prediction}
\label{subsec: vfi}
Our prediction module is built on a state-of-the-art imitation learning method~\cite{pari2021surprising}.
It consists of two phases: offline observation encoding (Line 1-5 of Alg. \ref{alg:method}) and online rollout (Line 6-17).
In the offline phase, we leverage a pre-trained vision encoder to project the observation image of every frame in $D$ to a deep embedding $\mathbf{z}^{e} \in Z^{e}$ (Line 3-5).
The embedding holds a compact representation for the visual observation.
We use a recently proposed self-supervised visual representation model~\cite{ibot} as the pre-trained visual encoder and compare the performance of different representation models in Sec. \ref{sec:expr}.

During the online rollout phase, at each time step $t$, the captured observation image $\mathbf{o}_{t}$ is also encoded as $\mathbf{z}_{t}$ using the same visual encoder (Line 10).
$\mathbf{z}_{t}$ is used to predict the action and target wrench by comparing with all the embeddings in $Z^{e}$ (Line 11).
Specifically, we use the cosine distance to compute the similarity between $\mathbf{z}_{t}$ and each embedding in $Z^{e}$:
\begin{equation}
    \mathrm{sim}(\mathbf{z}_{t}, \mathbf{z}_{i}^{e}) = \frac{{\mathbf{z}_{t}}^{T}{\mathbf{z}^{e}_{i}}}{\lVert \mathbf{z}^{e}_{i}\rVert \lVert \mathbf{z}_{t} \rVert}
\end{equation}
where $\mathbf{z}^{e}_{i}$ is the embedding of the i-th frame in $D$. 
The top-1 frame index in $D$ is retrieved via maximum similarity:
\begin{equation}
    i^{*} = \argmax_{i} \mathrm{sim}(\mathbf{z}_{t}, \mathbf{z}^{e}_{i})
\end{equation}
The kinematic action, gripper action, and terminate flag of the top-1 frame are used as the predicted kinematic action, gripper action and terminate flag: $\mathbf{a}_{t} = \mathbf{a}^{e}_{i^{*}}$, $\mathbf{g}_{t} = \mathbf{g}^{e}_{i^{*}}$, $\mathcal{T}_{t} = \mathcal{T}^{e}_{i^{*}}$.
The wrench of the \textit{next} frame of the top-1 frame is used as the target wrench for the next time step $\widehat{\mathcal{F}}_{t+1} = \mathcal{F}_{i^{*}+1}^{e}$.

\subsection{Admittance Whole-Body Control}
\label{subsec: awbc}
The current pose of the robot end-effector is denoted as $\mathcal{P}_{t} = [R_{t}, \mathbf{p}_{t}]\in SE(3)$ where $R_{t} \in SO(3)$ and $\mathbf{p}_{t} \in \mathbb{R}^{3}$ denote the rotation and translation, respectively.
The target end-effector pose for the next time step $\widehat{\mathcal{P}}_{t+1} \in SE(3)$ is computed by $\widehat{\mathcal{P}}_{t+1} = \mathcal{P}_{t} \circ \mathbf{a}_{t}$ where $\circ$ is the group action of $SE(3)$ (Line 14).
However, this pose may not be accurate due to the uncertainties caused by localization and the insufficient accuracy caused by action prediction (Sec. \ref{sec:intro}).
And small pose errors in contact-rich tasks may lead to large contact wrenches and even mechanical damage.
Therefore, we leverage an admittance control scheme and control the robot to track a pose that augments the original $\widehat{\mathcal{P}}_{t+1}$ with an admittance term $\Delta \mathcal{P}_{t+1} = (\Delta R_{t+1}, \Delta \mathbf{p}_{t+1}) \in SE(3)$ (Line 15-16).
This term compensates for the difference between the wrench at the current time step measured by the force-torque sensor $\mathcal{F}_{t}$ and the target wrench $\widehat{\mathcal{F}}_{t+1}$.
It is computed based on the wrench tracking error:
\begin{equation}
\begin{bmatrix}
\Delta \mathbf{w}_{t+1}\\
\Delta \mathbf{p}_{t+1}
\end{bmatrix} = \boldsymbol{\mathrm{K}}_{p}{\Delta{\mathcal{F}}}
    + \boldsymbol{\mathrm{K}}_{i}\int{\Delta{\mathcal{F}} \mathrm{d}t}
    + \boldsymbol{\mathrm{K}}_{d}{\frac{\partial \Delta{\mathcal{F}}}{\partial t}}
\label{eq: pid}
\end{equation}
where $\Delta R_{t+1}=\exp(\widehat{\Delta \mathbf{w}_{t+1}})$;
$\Delta{\mathcal{F}} = \widehat{\mathcal{F}}_{t+1} - \mathcal{F}_t$;
$\boldsymbol{\mathrm{K}}_{p}$, $\boldsymbol{\mathrm{K}}_{i}$ and $\boldsymbol{\mathrm{{K}}}_{d}$ are the proportional, integral, and differential terms, respectively. 
To obtain the final $\Delta \mathcal{P}_{t+1}$ applied to the robot, we discard the component parallel to $\mathbf{a}_{t}$ and only retain the perpendicular component.
This is equivalent to force-position hybrid control in which the position weight and the force weight are set to: 1 and 0 in the desired motion directions; 0 and 1 in the remaining directions.

With the admittance compensation, the augmented target pose $\widehat{\mathcal{P}}_{t+1}$ is sent to a whole-body controller to generate control commands for the mobile manipulator.
The whole-body controller is expressed as:
\begin{equation}
\begin{aligned}
    \min_{\mathbf{u}} \quad & f = \frac{1}{2} \mathbf{u}^T \mathbf{Q}\mathbf{u} + \mathbf{c}^T\mathbf{u} \\
    \mathrm{s.t.} \quad 
    &\mathbf{J}\mathbf{u} = \mathbf{v}_e\\
    &\mathbf{A} \mathbf{u} \le \mathbf{B}\\ 
    &\mathbf{u} \in [\mathbf{u}_{\mathrm{min}}, \mathbf{u}_{\mathrm{max}}]\\
\end{aligned}
\end{equation}
$\mathbf{u}$ is the decision variable vector which includes velocity control for the base and arm. 
$\mathbf{Q}$ incorporates joint velocity costs.
$\mathbf{c} = (\mathbf{0}_b,\;\mathbf{J}_m^a)$ tries to maximize the manipulability of the arm, where $\mathbf{0}_b$ is a zero vector and $\mathbf{J}_m^a$ is the manipulability Jacobian of the arm \cite{haviland2021neo}.
$\mathbf{J}$ is the generalized Jacobian of the base and arm.
$\mathbf{v}_e \in \mathbb{R}^{6}$ is the desired spatial velocity calculated by $\widehat{\mathcal{P}}_{t+1}$ and the current end-effector pose.
$\mathbf{A}$ and $\mathbf{B}$ implement the joint position constraints.
The optimization problem is a QP problem with variables $\mathbf{u}$.
We use qpOASES \cite{qpOASES} to solve the optimization problem in this paper.

%% file: 40experiment.tex
\begin{figure}[t]
\centering
\includegraphics[width=0.7\columnwidth]{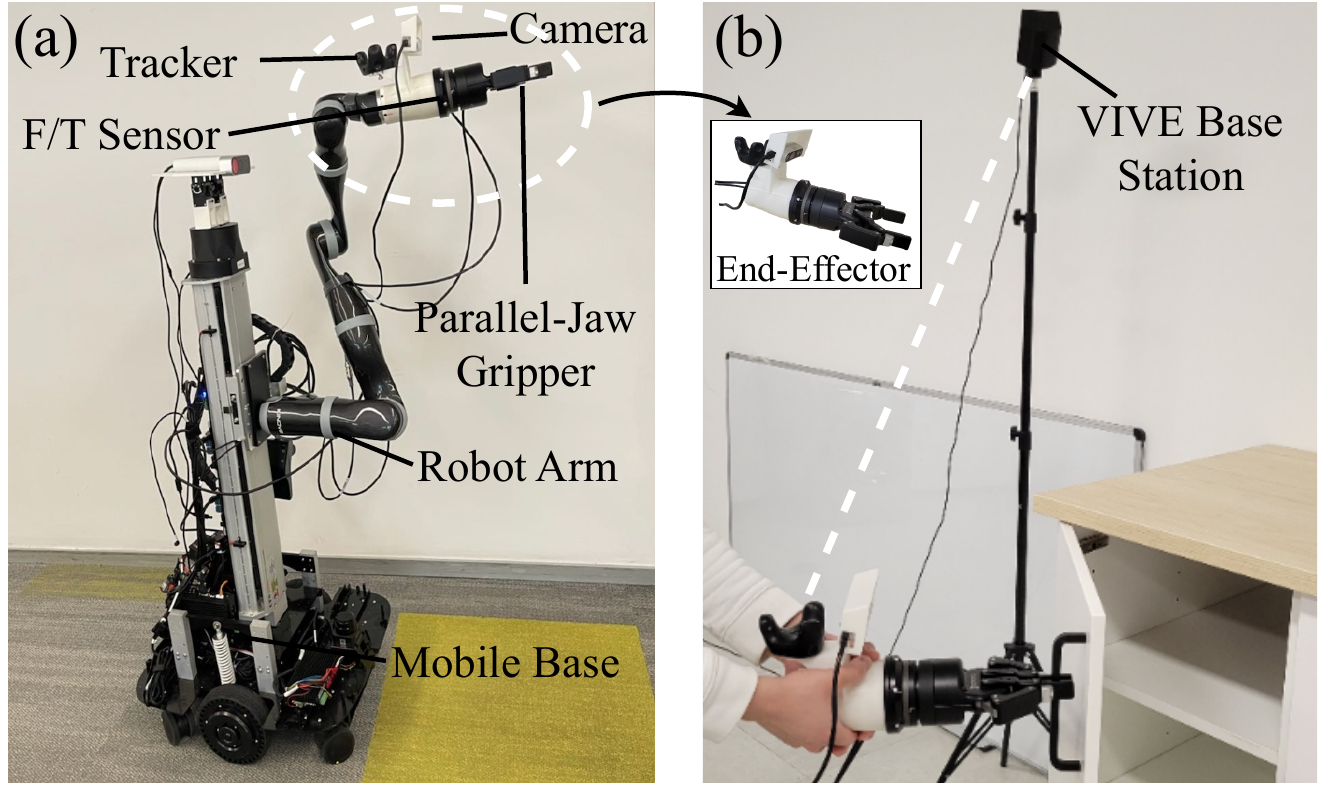}
\caption{\textbf{Experiment.} (a) The mobile manipulator used in the experiment. (b) Expert trajectory collection. \label{fig:hardware}}
\vspace{-0.5cm}
\end{figure}

\begin{figure*}
\centering
\includegraphics[width=0.95\linewidth]{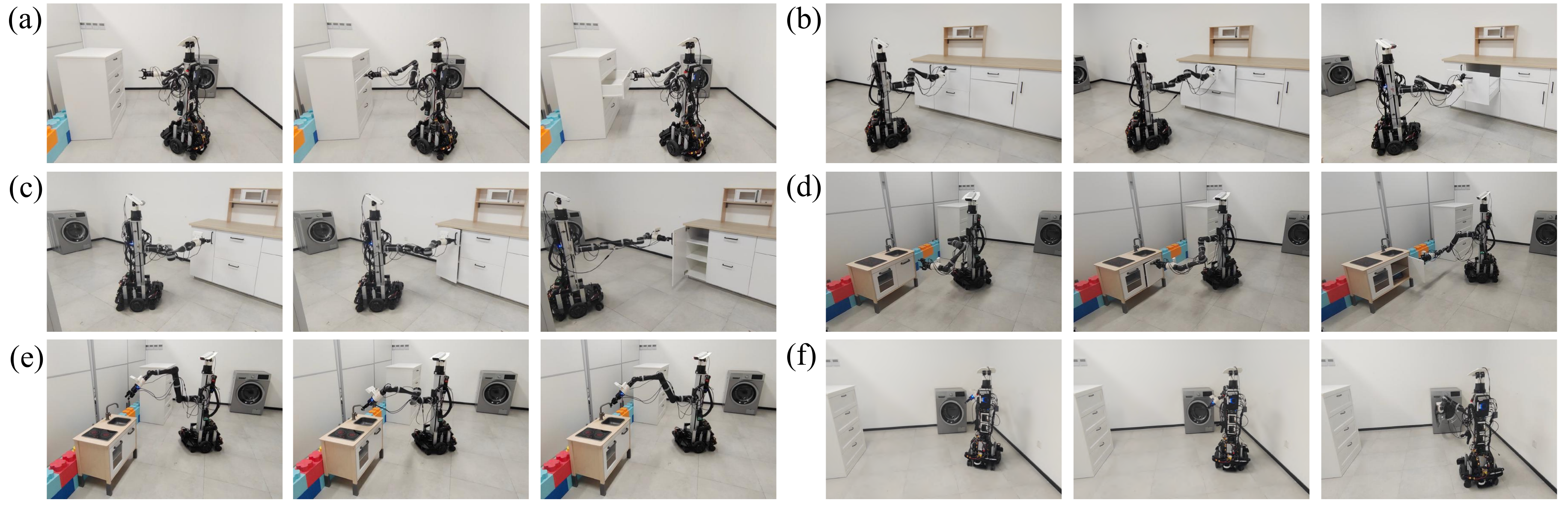}
\caption{\textbf{Real Robot Experiments.} 
(a) Open cabinet drawer. (b) Open drawer. (c) Open left door. (d) Open right door. (e) Rotate tap. (f) Open washing machine.}
\label{fig:experiment}
\vspace{-0.3cm}
\end{figure*}

We perform extensive real robot experiments on six contact-rich tasks (Fig. \ref{fig:illu}).
Through the experiments, we aim to answer three questions. 1) Is force imitation effective on improving task success rate? 2) Does force imitation effectively reduce undesired contact wrenches and unstable oscillations between the robot and the environment? 3) How do different visual encoders perform in terms of action prediction on real robot data?

\subsection{Experiment Setup} 
\label{subsec:experiment_setup}
\subsubsection{System}
The robot platform is shown in Fig. \ref{fig:hardware}(a). 
It consists of a differential-drive mobile base and a 7-DOF Kinova Gen2 robot arm. 
Wheel encoder, IMU, and lidar are used for the localization of the mobile base.
A Robotiq 2F-85 parallel-jaw gripper and a Robotiq FT 300-S force-torque sensor are mounted on the end-effector of the arm. 

\newcolumntype{Z}{p{0.8cm}<{\centering}}
\newcolumntype{W}{p{1.5cm}<{\centering}}
\newcolumntype{X}{p{1.7cm}<{\centering}}
\begin{table*}[t]
\centering
\caption{Success Rates of Real Robot Experiments.}
\begin{tabular}{c|XWWWWX|c}
\toprule
{Method}&{Open cabinet drawer}&{Open drawer}&{Open left door}&{Open right door}&{Rotate tap}&{Open washing machine}&{Average} \\
\hline
{BC (single-task) \cite{nair2022r3m}}&{0\%}&{20\%}&{40\%}&{0\%}&{50\%}&{10\%}&{20.0\%}\\
{\ourmethod w/o FC \& Rot}&{20\%}&{50\%}&{10\%}&{0\%}&{20\%}&{0\%}&{16.7\%}\\
{\ourmethod w/o FC}&{70\%}&{70\%}&{50\%}&{60\%}&{10\%}&{10\%}&{45.0\%}\\
\hline
{\ourmethod (Ours)}&{70\%}&{70\%}&{80\%}&{60\%}&{60\%}&{100\%}&{\textbf{73.3\%}}\\
\bottomrule
\end{tabular}
\label{tab:benchmark}
\vspace{-0.4cm}
\end{table*}

\subsubsection{Tasks}
We perform experiments on six contact-rich tasks in which the robot manipulates different articulated objects (Fig. \ref{fig:illu}).
The articulation includes prismatic joints (\textit{e.g.}, drawers) and revolute joints (\textit{e.g.}, doors and the tap).
\textit{Almost all tasks are impossible to finish by static manipulations}.
All tasks require the robot to grasp the object and stay in contact with it during manipulation.

\subsubsection{Expert Data Collection}
We collected 30 expert demonstrations for each task.
The human expert carried the end effector to collect expert trajectories (Fig. \ref{fig:hardware}(b)).
The observation RGB image, the pose of the end effector, the gripper action, and the wrench were recorded for each time step.
The pose of the end effector is tracked via an HTC VIVE tracking system.
The kinematic action for each frame is computed as the relative transformation between the pose of the next time step and the pose of the current time step.
The frames in which experts perform gripper actions have no kinematic actions.
That is, if this frame has been retrieved in the rollout, the robot executes the gripper action and the pose of the robot remains unchanged.
The terminate flag is 1 for the last frame and 0 otherwise.

\subsubsection{Pipeline}
All tasks can be divided into three phases: approaching, grasping, and in-contact manipulation.
The robot starts at a pre-defined initial pose which is contact-free.
We randomize the end-effector initial pose before a rollout starts.
The robot first performs predicted actions to approach the object.
When the robot is commanded to close the gripper, the robot enters the grasping phase.
Finally, the robot enters the in-contact manipulation phase.
A rollout terminates if either one of the three conditions is met: 1) the termination flag $\mathcal{T}_{t} = 1$; 2) the rollout time step exceeds the maximum time step; 3) the force is larger than 40N or the average force in the past 1s is larger than 30N.
The last condition ensures safety and avoids mechanical damage to the object and the robot during rollouts.
We consider a trial successful if at least 80\% of the task has been completed.
Each task is carried out 10 times for each method.

\subsection{Experiment Results} 
We denote our visual-force imitation method as \texttt{\ourmethod}.
We compare with three baseline methods.
The first baseline method is behavior cloning, denoted as \texttt{BC}.
We use the same visual encoder \cite{ibot} to encode observation images to representation vectors.
We follow \cite{nair2022r3m} and train a multi-layer perceptron that takes as input the representation vector and outputs the kinematic action, gripper action and terminate flag.
In particular, we train \texttt{BC} in a single-task learning setting.
We performed multi-task learning but found the results not good.
We also compare with a variant of our method without force imitation, denoted as \texttt{\ourmethod w/o FC}, to validate the effectiveness of force imitation.
In addition, we test a variant of our method without force imitation and rotation, \textit{i.e.}, the action only contains translation $\mathbf{a}_{t} \in \mathbb{R}^{3}$.
We denote this method as \texttt{\ourmethod w/o FC \& Rot}.
Qualitative and quantitative results are shown in Fig. \ref{fig:experiment} and Tab. \ref{tab:benchmark}, respectively.

\textit{1) \textbf{Is force imitation effective in improving task success rate?}}
Our method \texttt{\ourmethod} achieves the best average success rate among all the comparing baseline methods.
Compared to \texttt{BC}, our method obtains a performance gain of 53.33\% even though it performs multi-task learning while \texttt{BC} performs single-task learning.
\texttt{BC} struggles with the compounding-error problem when approaching the object in a few tasks.
The grasping success rates in these tasks are low.
In other tasks where grasping is mostly successful, it cannot maintain a small force, resulting in early termination without task completion.
Without force imitation, the success rate of \texttt{\ourmethod w/o FC} decreases by 28.3\% on average.
This method mainly struggles with tasks involving translation and rotation, \textit{e.g.}, opening the washing machine and rotating the tap.
In these tasks, similar to \texttt{BC} which also does not have force imitation, it suffers from early termination due to large forces.

\begin{figure}[htbp]
\centering
\includegraphics[width=0.85\columnwidth]{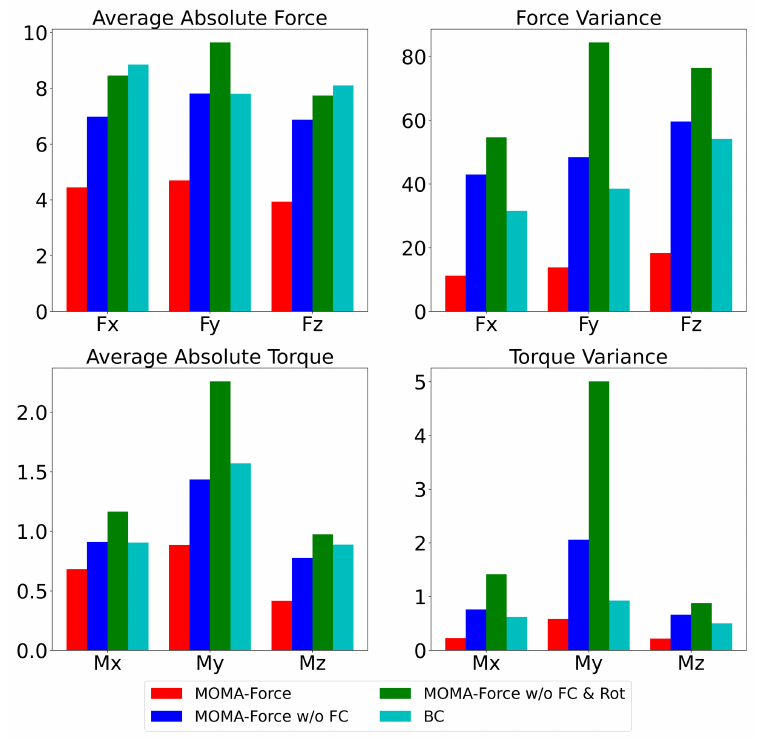}
\caption{\textbf{Results of Contact Wrenches.} The force imitation results in a reduction on contact wrench (N and Nm) and variance (N$^2$ and N$^2$m$^2$).
\label{fig:wrench}}
\vspace{-0.3cm}
\end{figure}

The \textbf{failure cases} of \texttt{\ourmethod} mainly attribute to inaccurate action prediction.
A typical failure mode is unsuccessful grasping.
In the expert trajectories, the observation images between frames with small pose changes are very similar.
Therefore, in some cases, the visual encoder is not able to distinguish and thus the robot cannot retrieve the best expert action.

\textit{2) \textbf{Is force imitation effective on decreasing the contact wrench between the robot and the environment?}}
We compare the contact wrench of \texttt{\ourmethod} and the baseline methods. For all methods, we first compute the average absolute contact wrenches and average wrench variances of all the trials in which the grasping is successfully in the rollout for each of the six tasks.
We then average across tasks and report the results in Fig. \ref{fig:wrench}.
Lower variances indicate more stable contact.
With force imitation, the average absolute contact force and torque of \texttt{\ourmethod} in x, y, and z-axes are all smaller compared to those of the baseline methods without force imitation.
In addition, \texttt{\ourmethod} has a smaller force variance, indicating less oscillation and more stable contact during the rollout.

\begin{table}[t]
\centering
\caption{Mean Squared Error of predicted actions and wrenches. \label{tab:visual}}
\begin{tabularx}{\linewidth}{X<{\centering}|WWW}
\toprule
{Visual Encoder}&{Translation}&{Rotation}&{Wrench} \\
\hline
{MVP \cite{radosavovicreal}}&{0.19}&{0.17}&{0.21}\\
{CLIP \cite{radford2021learning}}&{0.16}&{0.17}&{0.15}\\
{iBOT \cite{ibot}}&{\textbf{0.10}}&{\textbf{0.11}}&{\textbf{0.10}}\\
\bottomrule
\end{tabularx}
\vspace{-0.5cm}
\end{table}

\textit{3) \textbf{How do different pre-trained visual encoders perform on real-robot data?}}
To compare the effectiveness of different visual pre-trained models, we show the Mean Squared Error (MSE) of using various state-of-the-art pre-trained models as the visual encoder on the test set with 5-fold cross-validation.
Tab. \ref{tab:visual} shows the result.
MVP \cite{radosavovicreal} is pre-trained via a masked autoencoder from the Internet and egocentric videos. 
CLIP \cite{radford2021learning} aims to align the image representation with the paired text through contrastive learning. 
iBOT \cite{ibot} gives a good trade-off between masked autoencoder and contrastive learning with an online tokenizer. 
Tab. \ref{tab:visual} shows that iBOT achieves the best performance, 
demonstrating its advantages on our tasks. 
We note that a concurrent work~\cite{siddharth2023language} gives a more sufficient study on visual representations for robot tasks.

%% file: 60conclusion.tex
In this work, we present a novel visual-force imitation learning method for real-world, contact-rich mobile manipulation tasks.
We extend a state-of-the-art visual imitation learning method to support both kinematic action and target wrench prediction.
We leverage admittance whole-body control to enable robots to track the trajectory generated by the action while regulating the wrench to follow the expert wrench.
We implement our method on a high DoF mobile manipulator and perform multi-task learning on six real-world contact-rich tasks.
Our method achieves an average success rate of 73.3\%, outperforming several baseline methods without force imitation.
In addition, the average absolute contact wrench and wrench variance of our method are smaller compared to all the baseline methods.

%% file: appendix.tex
\subsection{Expert Data Details}
For each task, we collect 30 expert demonstrations as described in Sec. \ref{subsec:experiment_setup}.
For each expert demonstration, we first filtered out frames that are static.
Following~\cite{bcz}, the expert action of a frame is computed as the state difference between the frame and its next $N$-th frame.
We set $N=1$ first and then increment $N$ such that either one of the conditions is met: 1) the translation difference is larger than 1cm; 2) the rotation difference is larger than 0.1rad; 3) the gripper status changes.

\subsection{Experiment Details}
\begin{figure}[h]
    \centering
     \includegraphics[width=1\columnwidth]{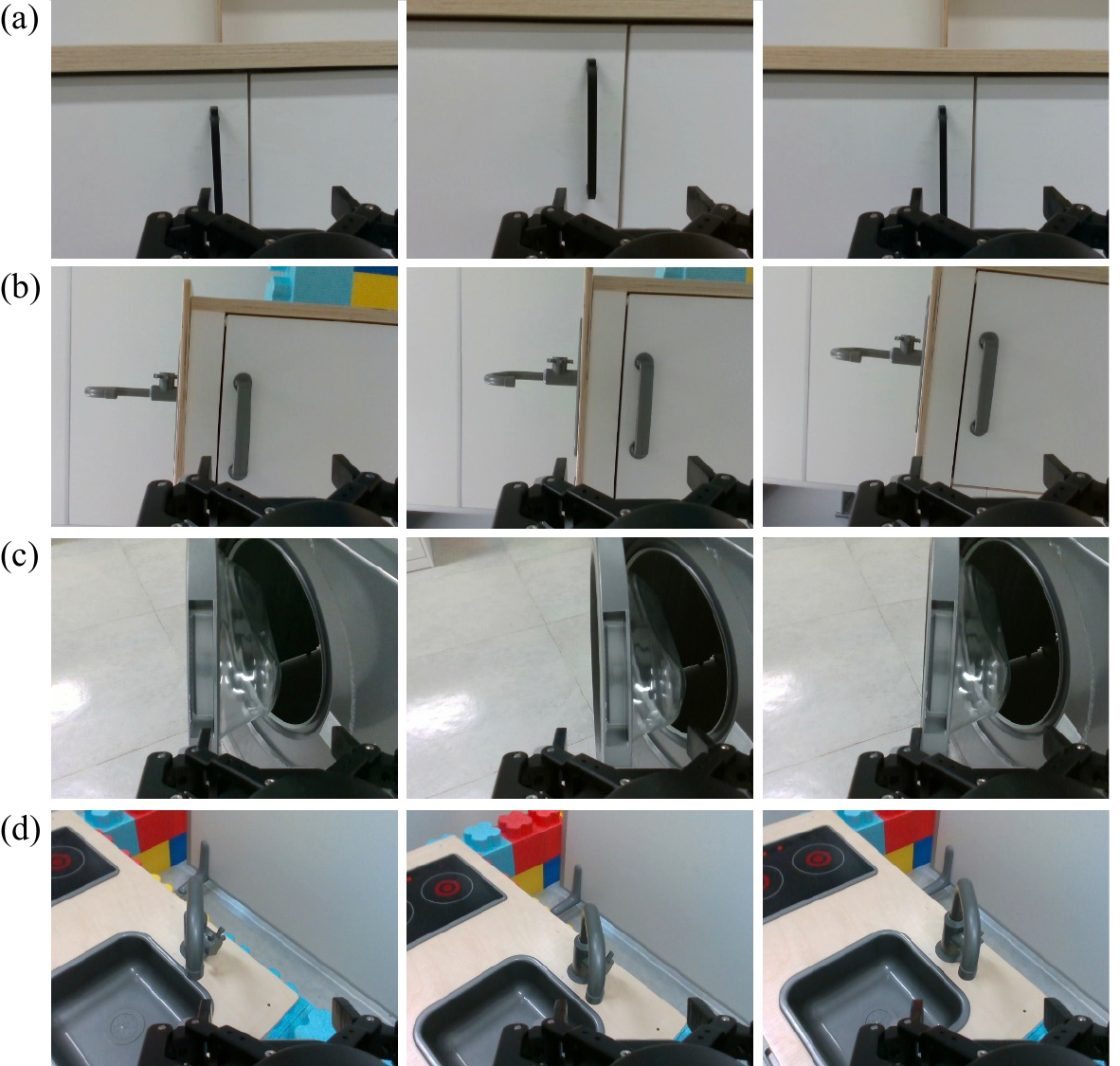}
    \caption{\textbf{Initialization.} We show the initialize observation images from three trials of (a) open left door, (b) open right door, (c) open washing machine, and (d) rotate tap.}
    \label{fig:appendix_initialization}
\end{figure}

For \ourmethod and its variants, the maximum rollout iteration is 200; for BC, the maximum iteration is 250.
For each trial, the robot is first controlled to move its end-effector to a pre-defined initial pose.
We then randomize the pose with a uniform noise $(\delta x, \delta y, \delta z, \delta \alpha, \delta \beta, \delta \gamma)$ in which  $\delta x$, $\delta y$, and $\delta z$ are sampled from $U(-2, 2)$cm; $\delta \alpha$, $\delta \beta$, and $\delta \gamma$ are sampled from $U(-0.06, 0.06)$rad (Fig. \ref{fig:appendix_initialization}).
$\alpha$, $\beta$, and $\gamma$ are the Euler angles.
For all the tasks, the robot motion can be divided into three phases: approaching, grasping, and in-contact manipulation.
In the approaching phase, as the robot is close to the object, the mobile base is locked and the robot only performs arm motion.
After the robot grasps the object, force control is activated and the base is unlocked, \textit{i.e.}, the robot enters whole-body control mode.
We use PID control to regulate the wrench to the target wrench.
The PID parameters are shown in Tab. \ref{tab:pid}.
The derivative term ${\partial \Delta{\mathcal{F}} / {\partial t}}$ in Eq. \ref{eq: pid} is determined by the difference between the wrench errors $\Delta{\mathcal{F}}$ in two consecutive time steps. 

\begin{table}[t]
\centering
\caption{PID Parameters for Force Control.}
    \begin{tabular}{c | c c c c c c}
        \toprule
        {Terms}&{$F_x$}&{$F_y$}&{$F_z$}&{$M_x$}&{$M_y$}&{$M_z$} \\
        \hline
        $K_{p}$ &{4e-4}&{4e-4}&{4e-4}&{4e-4}&{4e-4}&{4e-4}\\
        $K_{i}$ &{5e-5}&{5e-5}&{5e-5}&{5e-5}&{5e-5}&{5e-5}\\
        $K_{d}$ &{5e-4}&{5e-4}&{5e-4}&{5e-4}&{5e-4}&{5e-4}\\
        \bottomrule
    \end{tabular}
\label{tab:pid}
\end{table}

\subsection{Future Work}
For future work, we plan to explore more on extending our method to more complex scenarios including perturbed backgrounds and novel objects.
In addition, we find in the experiment that grasping is important in terms of overall task success rate and force control.
While we use a parallel-jaw gripper in this work, future work can also investigate using more sophisticated grippers (\textit{e.g.}, robot hands) which can be actively controlled during the course of rollouts.